%% file: neurips_2024.tex
\documentclass{article}

\PassOptionsToPackage{numbers, compress}{natbib}

\usepackage[preprint]{neurips_2024}

\usepackage[utf8]{inputenc} 
\usepackage[T1]{fontenc}    
\usepackage{hyperref}       
\usepackage{url}            
\usepackage{booktabs}       
\usepackage{amsfonts}       
\usepackage{nicefrac}       
\usepackage{microtype}      
\usepackage{xcolor}         
\usepackage{graphicx}
\newtheorem{thm}{Theorem}
\newtheorem{definition}[thm]{Definition}

\usepackage{amsmath}
\usepackage{multirow}
\usepackage{amssymb}
\usepackage{algorithm}
\usepackage{algorithmic}
\usepackage{wrapfig}
\usepackage{tabularx} 
\usepackage{multirow} 
\usepackage{booktabs} 
\usepackage{units}
\usepackage{dirtytalk}

\title{Denoising Reuse: Exploiting Inter-frame Motion Consistency for Efficient Video Latent Generation}

\DeclareMathOperator{\argmin}{argmin}

\author{%
Chenyu Wang \quad Shuo Yan \quad Yixuan Chen \quad Yujiang Wang \quad Mingzhi Dong \\
\textbf{Xiaochen Yang} \quad \textbf{Dongsheng Li} \quad \textbf{Robert P. Dick} \quad \textbf{Qin Lv} \quad \textbf{Fan Yang} \\
\quad \textbf{Tun Lu} \quad \textbf{Ning Gu} \quad \textbf{Li Shang}\\
}

\begin{document}

\maketitle





\begin{abstract}
\label{sec:abstract}

Video generation using diffusion-based models is constrained by high computational costs due to the frame-wise iterative diffusion process. This work presents a Diffusion Reuse MOtion (Dr. Mo) network to accelerate latent video generation.
Our key discovery is that coarse-grained noises in earlier denoising steps have demonstrated high motion consistency across consecutive video frames. 
Following this observation, Dr. Mo propagates those coarse-grained noises onto the next frame by incorporating carefully designed, lightweight inter-frame motions, eliminating massive computational redundancy in frame-wise diffusion models.
The more sensitive and fine-grained noises are still acquired via later denoising steps, which can be essential to retain visual qualities.
As such, deciding which intermediate steps should switch from motion-based propagations to denoising can be a crucial problem and a key tradeoff between efficiency and quality. 
Dr. Mo employs a meta-network named Denoising Step Selector (DSS) to dynamically determine desirable intermediate steps across video frames.
Extensive evaluations on video generation and editing tasks have shown that Dr. Mo can substantially accelerate diffusion models in video tasks with improved visual qualities. 
\end{abstract}

\input{1_intro}
\input{2_analysis}

\input{3_method}

\input{5_experiment}
\input{4_related}

\section{Conclusion}

This paper addresses the efficiency challenges in diffusion-based video generation methods, inspired by a key observation that inter-frame motion features remain consistent through most of the diffusion process. The proposed method, called Dr.~Mo, enables the reuse of frames across multiple denoising steps, which significantly reduces the need to regenerate each frame from scratch, thereby lowering the computational load and speeding up the video generation process. Frame-specific updates are applied only in the final stages of denoising to maintain the video's integrity and detail. Evaluations in video generation and editing show that Dr.~Mo increases the speed of video generation by a factor of 4 compared to Latentshift, and 1.5 times compared to SimDA and LaVie. Our future work aims to enhance video generation of visually rich features with complex motion transformations.


\newpage
\bibliography{neurips_2024}
\bibliographystyle{neurips_2024}

\newpage
\section*{Appendix}

\appendix
\input{0_appendix}

\end{document}

%% file: 1_intro.tex
\section{Introduction}
\label{sec:intro}


Diffusion models such as Denoising Diffusion Probabilistic Models (DDPMs)~\citep{ho2020denoising} and Video Diffusion Models (VDMs)~\citep{ho2022video} have demonstrated impressive capabilities to generate high-fidelity videos from still images that suggest the desired style and content. 
However, the superior visual qualities come at the cost of computation burdens primarily associated with the iterative diffusion process, which consists of multiple denoising steps \citep{ronneberger2015u, song2020denoising}.
This is cost prohibitive for videos; frame-wise application of diffusion models imposes computational demands that increase linearly with the number of frames, undermining the generation of long-duration videos \citep{ho2022video}.

\begin{figure}[t]
    \centering
    \includegraphics[width=0.8\textwidth]{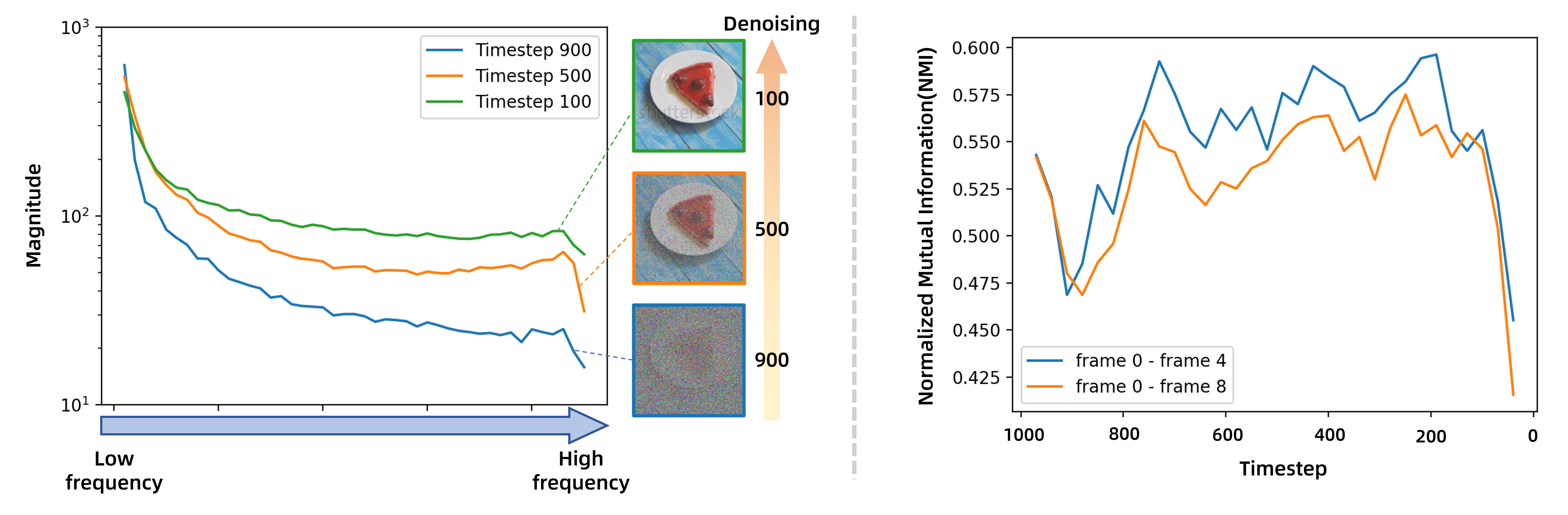}
    \caption{Left: The spectrum illustrates an increase in high-frequency signals during the denoising process, from steps 900 to 100. Right: High NMI scores between steps 800 and 200 indicate consistent motion dynamics of video frames 0-4 (and 0-8) throughout the denoising process.}
    \label{fig:intro_figure1}
\end{figure}

This work aims to dramatically accelerate diffusion-based video generation by using motion dynamics in the latent space.
We first delve into the video generation process to illustrate our insights.
As shown in Figure~\ref{fig:intro_figure1} (left), the diffusion model applies incremental noise reduction to gradually produce visual features of better qualities and higher resolutions, reflecting coarse- to fine-grained patterns.
We subsequently analyze the inter-frame motion dynamics throughout the denoising phase \footnote[1]{These dynamic changes are quantified by the normalized mutual information (NMI) between learned motion matrices. Higher NMI indicates better consistency. Details are provided in  Section~\ref{sec:empirical_consist}.}.
Figure~\ref{fig:intro_figure1} (right) shows that inter-frame motion features are consistent across many the denoising steps, especially those operating on coarse-grained features.
This reveals an way to accelerate diffusion-based video generation: latent residuals in one video frame can be reused to rapidly estimate those in subsequent frames.

Residuals in later steps, however, cannot be similarly estimated: they are more directly and linked to the generated images and require precision to maintain the desired visual quality. 
Thus, it is possible to dramatically accelerate the denoising process in video generation, but only if the appropriate (and inappropriate) denoising steps for efficient, motion-based residual estimation can be determiend.
Transitioning from motion-based estimation to early undermines efficiency; transitioning too late undermines quality.

We describe a new Diffusion Reuse MOtion (Dr.~Mo) network that accelerates the frame-wise diffusion models using inter-frame motion for efficient estimation of latent residuals.
Dr.~Mo first applies a diffusion model to a frame image to obtain step-wise residuals: the base latent representation.
Motion matrices are constructed to capture semantic motion features across video frames, which are learned from the semantically rich visual features extracted by a U-Net-like decoder~\cite{ronneberger2015u}.
When generating a frame, Dr.~Mo uses a novel meta-network, the Denoising Step Selector (DSS), to determine the proper denoising step for transitioning away from motion-based residual estimation.
Latent residuals before the transition step are rapidly estimated using the motion matrices and base latent representations of the corresponding denoising step.
After the transition step, latent residuals are processed by the rest of the diffusion model and output to produce the final frame. 

We compare Dr.~Mo with state-of-the-art baselines on the UCF-101~\citep{soomro2012ucf101} and MSR-VTT~\citep{xu2016msr} datasets and demonstrate superior video quality and semantic alignment.
Notably, Dr.~Mo effectively accelerates the generation of 16-frame 256$\times$256 videos by a factor of 4 compared with Latent-Shift~\citep{an2023latentshift}, while maintaining 96\% of the IS~\cite{salimans2016improved} and achieving improved FVD~\cite{unterthiner2019fvd}.
Additionally, Dr.~Mo generates 16-frame 512x512 videos at 1.5 times the speed of SimDA~\citep{xing2023simda} and LaVie~\citep{wang2023lavie}. Furthermore, Dr.~Mo supports video style transfer by simply providing a style-transferred first frame.

In summary, our work makes the following contributions:
\begin{enumerate}

\item We find that motion information is consistent throughout most of the stable diffusion process, which facilitates easy learning and inter-frame transformations. 

\item We describe a lightweight motion learning module that efficiently captures and uses inter-frame motion features to accelerate video generation in diffusion models.

\item We design a meta-network to dynamically determine the reusable denoising steps enabling tradeoffs between video generation efficiency and quality. 

\end{enumerate}

Compared with prior work on video generation and editing, Dr.~Mo improves computational efficiency and video quality . 


%% file: 2_analysis.tex
\section{Motion Dynamics in Diffusion Model}
\label{sec:analysis}
This section analyzes motion dynamics \textcolor{black}{throughout the coarse- to fine-grained visual feature generation process}. We find that motion dynamics are consistent in the majority of denoising steps but that the optimal number of reuse steps is frame dependent. This  phenomenon motivates us to adaptively reuse denoising steps across frames for efficient video generation.

\begin{figure}[t]
    \centering
    \includegraphics[width=0.8\textwidth]{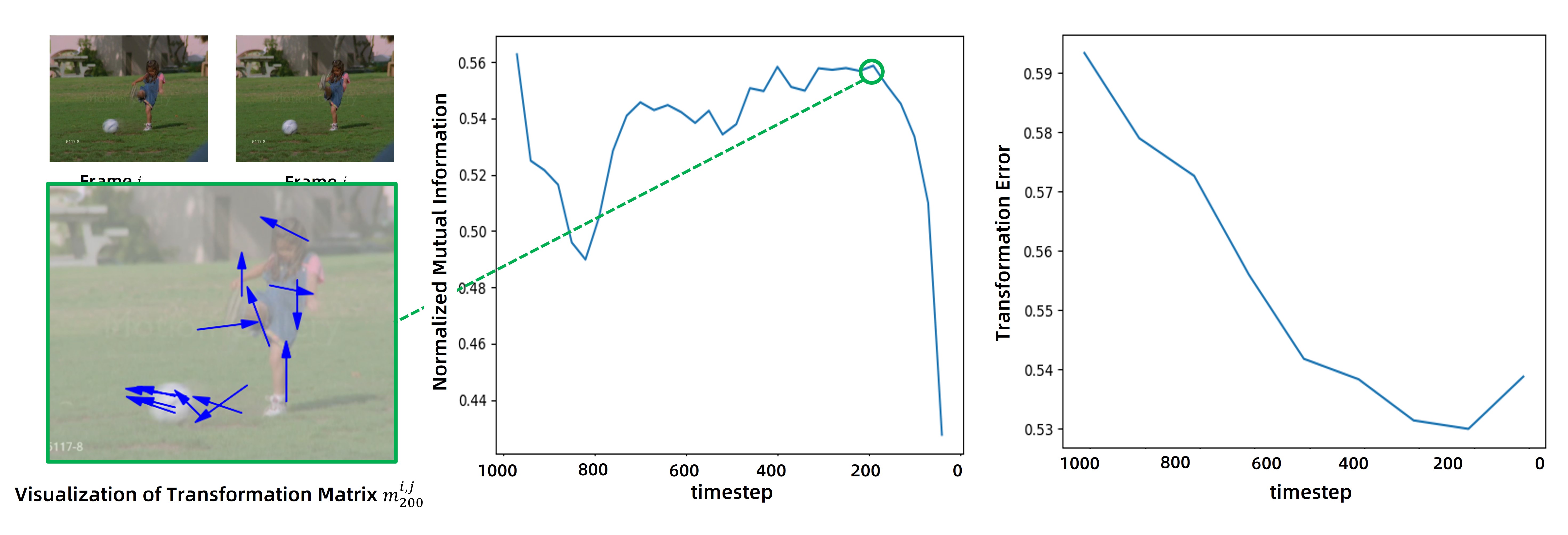}
    \caption{
    Motion visualization at step 200 accurately captures the movement trends of patch features. At this step, the motion dynamics show consistency with low transformation errors, indicating the potential for reusing steps between step  1000 and 200.
    }
    \label{fig:analysis}
\end{figure}

\subsection{Motion Dynamics}
\label{sec:transformation_operation_understanding}
In this study, we employ the Stable Diffusion (SD) model as our foundational diffusion model for generating videos. Consider a video comprising $F$ frames, denoted by $I = [I^1, \ldots, I^F]$. Initially, each frame $I^i$ is encoded into a latent space representation $\mathbf{z}^i$. We employ the DDPM approach with $T=1000$ denoising steps to recover the original frames. 
The denoising process recovers each frame from step $T$ to 1. $\mathbf{z}_t^i$ represents the latent state of frame $i$ at timestep $t$, where $t$ indicates the timestep and $i$ indicates the frame number within the video sequence.

To analyze the inter-frame motion dynamics for generating coherent videos using a diffusion model, we introduce the concept of latent residual to represent the change in latent features between two steps, denoted as: 
\begin{equation}
    \label{equ:residual}
    \delta \mathbf{z}_t^i := \mathbf{z}_{t-1}^i - \mathbf{z}_t^i.
\end{equation}
This difference can be regarded as the feature revealed (or noise removed) due to the denoising process. Consequently, the latent representation at step $t$ for frame $i$ can be reconstructed by summing the following residuals: $\mathbf{z}_t^i = \mathbf{z}_T^i + \sum_{k=t+1}^{T} \delta \mathbf{z}_k^i$, where $\mathbf{z}_T^i$ denotes the initial noisy image at the start of the reverse denoising process. 

Next, we introduce the concept of a transformation operation between frames (denoted as $g$) to characterize inter-frame motion dynamics in latent residuals corresponding to the same denoising step. Considering frames $i$ and $j$, $g_\phi^t$ transforms $\delta \mathbf{z}_t^i$ to match $\delta \mathbf{z}_t^j$ governed by minimizing the transformation error, as expressed by
\begin{equation}
    \label{equ:transformation_residual}
    \min_\phi \| \delta \mathbf{z}_t^j - g_\phi^t(\delta\mathbf{z}_t^i) \|_1.
\end{equation}

Drawing inspiration from optical flow techniques~\citep{horn1981determining}, we propose to learn motion dynamics between frames using function $\mathcal{C}(\cdot, \cdot)$ to generate motion matrix $\mathbf{M}_{\delta\mathbf{z}_t}^{i,j}$. 
This motion matrix describes the temporal relations between the residual $\delta \mathbf{z}_t^i$ and $\delta \mathbf{z}_t^j$ at the same denoising steps, defined as:
\begin{equation}
    \label{equ:transformation}
    g_\phi(\delta \mathbf{z}_{t}^i) = (\delta \mathbf{z}_{t}^i)^\top \times \mathbf{M}_{\delta \mathbf{z}_t}^{i,j}, \quad \text{where} \quad \mathbf{M}_{\delta \mathbf{z}_t}^{i,j} = \mathcal{C}(\delta \mathbf{z}_{t}^i, \delta \mathbf{z}_{t}^j) = (\frac{\delta \mathbf{z}_{t}^i}{\|\delta \mathbf{z}_{t}^i\|}) \cdot (\frac{\delta \mathbf{z}_{t}^j}{\|\delta \mathbf{z}_{t}^j\|})^\top.
\end{equation}
\textcolor{black}{
Here, $\mathcal{C}(\cdot, \cdot)$ denotes a motion modeling function based on the cosine-similarity computation~\citep{zadaianchuk2024object,zhong2023mmvp}, $\mathbf{M}_{\delta\mathbf{z}_t}^{i,j}$ can be regarded as a heatmap, indicating the moving transition probabilities between latent features. Details are provided in Section~\ref{sec:motion_learning}.}

\subsection{Temporal Consistency of Latent Motion Dynamics} 
\label{sec:empirical_consist}
This subsection defines and quantifies the temporal consistency of latent motion dynamics.

\begin{definition}[Step-wise Temporal Consistency of Motion Dynamics]
\label{def:consistency}
Given motion matrices $\mathbf{M}_{\delta \mathbf{z}_t}^{i,j}$ and $\mathbf{M}_{\delta \mathbf{z}_{t+1}}^{i,j}$ between frames $i$ and $j$ at timestep $t$ and $t+1$, the temporal consistency of motion dynamics is defined as the degree of similarity between the two matrices.
\end{definition}

To quantify this consistency, we use Normalized Mutual Information (NMI), defined as:
\begin{equation}
\label{equ:consistency}
\text{NMI}(\mathbf{M}_{\delta \mathbf{z}_{t}}^{i,j}, \mathbf{M}_{\delta \mathbf{z}_{t+1}}^{i,j}) = \frac{I(\mathbf{M}_{\delta \mathbf{z}_t}^{i,j}; \mathbf{M}_{\delta \mathbf{z}_{t+1}}^{i,j})}{\sqrt{H(\mathbf{M}_{\delta \mathbf{z}_t}^{i,j})}\sqrt{H(\mathbf{M}_{\delta \mathbf{z}_{t+1}}^{i,j})}},
\end{equation}
where $\mathbf{M}_{\delta \mathbf{z}_{t}}^{i,j}$ and $\mathbf{M}_{\delta \mathbf{z}_{t+1}}^{i,j}$ are motion matrices between frames $i$ and $j$ at timestep $t$ and $t+1$, respectively. 
$I$ represents mutual information and $H$ denotes entropy. By measuring the mutual information between motion matrices at different timesteps, NMI quantifies the predictive information about $\mathbf{M}_{\delta \mathbf{z}_{t+1}}^{i,j}$ from $\mathbf{M}_{\delta \mathbf{z}_{t}}^{i,j}$. Thus, high NMI values indicate a strong consistency of motion dynamics.



As illustrated in Figure~\ref{fig:analysis}, motion  consistency exists throughout most steps of the diffusion process. Specifically, from the beginning to 0.2T, i.e., 80\% of the denoising process, the data exhibits high NMI values and a decline in transformation errors, indicating consistent and reliable motion predictions. This consistency primarily stems from the presence of coarse-grained, semantically rich latent features that enhance the modeling of motion dynamics. 
In contrast, in the late denoising steps, e.g., from 0.2T to 0T, or the rest 20\% of the denoising process, the emergence of finer details increases the visual feature complexity, resulting in lower NMI and decreased predictability. 
These findings demonstrate the potential for reusing denoising steps across frames, which significantly enhances computational efficiency and accelerates video generation. 
\textcolor{black}{Moreover, it allows simple control over the tradeoffs between  efficiency and quality. }


%% file: 3_method.tex
\section{Dr.~Mo: Denoising Reuse for Efficient Video Generation}
This section presents Dr.~Mo, 
a diffusion reuse motion network that captures and uses inter-frame motion features to accelerate video latent generation in diffusion models.

\subsection{Overview}
\label{sec:overview}

Dr.~Mo consists of two main components: the -Motion Transformation Network (MTN) and Denoising Step Selector (DSS). The MTN develops step-specific motion matrices from residual latents and provides the motion sequence with its consistency information to the DSS. The DSS then determines which
intermediate step (denoted as $t^\ast \in [T]$) should switch from motion-based propagations to denoising in order to optimize the balance between computation efficiency and output quality. With the switching step determined, the MTN refines the final motion matrix for inter-frame transformations, enhancing the system's efficiency and video quality.

During inference, Dr.~Mo extracts motion matrices across various timesteps from two reference frames. These matrices are analyzed by the DSS to select the most suitable $t^\ast$. Using this selected switch step, the MTN extracts the motion matrix from reference frames at time $t^\ast$ and predicts the future sequence of motion matrices, which are used to generate future frames. Each frame undergoes a tailored denoising process from step $t^\ast$ to 1, ensuring optimized detail and visual integrity.

\subsection{Motion Transformation Network}
\label{sec:motion_learning}


\noindent \textbf{Motion Matrix Construction.} 
The outputs of U-Net represent the predicted noise to be removed from $\mathbf{z}_{t}$ to recover $\mathbf{z}_{t-1}$. Thus, the intermediate feature of U-Net provides estimates of the residuals between these steps. Furthermore, recent studies have demonstrated that intermediate diffusion features extracted from U-Net can capture coarse- and fine-grained semantic information ~\citep{kim2024leveraging,baranchuk2021label,liu2024drag, namekata2024emerdiff}. Therefore, we use the representations from the U-Net decoder to construct the motion matrix. 

Given two video frames $i$ and $j$, we extract features from multiple blocks $[b_1, \ldots, b_k]$ of the U-Net decoder at timestep $t$. 
Here, $b_{\cdot}$ represents the block index within the U-Net architecture. The features, denoted as $\delta \mathbf{z}_t^{i}[b_k]$ and $\delta \mathbf{z}_t^{j}[b_k]$, are processed through a convolutional network to generate block-specific motion matrices, which are then aggregated by a multi-layer perceptron (MLP) to construct a multi-scale motion matrix:
\begin{equation}
    \label{equ:combined_motion_calculation}
    \mathbf{M}_{\delta \mathbf{z}_t}^{i,j} = g_{\phi_2}([\mathbf{M}_{\delta \mathbf{z}_t}^{i,j}[b_1], \ldots, \mathbf{M}_{\delta \mathbf{z}_t}^{i,j}[b_k]]),\quad \text{where } \mathbf{M}_{\delta \mathbf{z}_t}^{i,j}[b_k] = \mathcal{C}(g_{\phi_1}(\delta \mathbf{z}_t^{i}[b_k]), g_{\phi_1}(\delta \mathbf{z}_t^{j}[b_k])),
    \end{equation}
where $\phi_1$ and $\phi_2$ denote the parameters of the convolutional network and the MLP, respectively. 
Since each block displays varying levels of semantic granularity, this leads to different motion dynamics. The computed motion matrix $\mathbf{M}_{\delta \mathbf{z}_{t}}^{i,j}$ captures the multi-scale motion dynamics within the residual latents. Further analysis can be found in the Supplementary Material.

\noindent\textbf{Motion Learning Objectives.}
The first learning objective is to minimize the transformation error between latent variables $\delta \mathbf{z}_t^i$ and $ \delta \mathbf{z}_t^j$ at each denoising step: 
\begin{equation}
    \label{equ:transformation_error_delta}
    \mathcal{L}^{\text{visual}}_{\delta\mathbf{z}} = \sum_{i,j,t}|| (\delta \mathbf{z}_t^{i})^\top \times \mathbf{M}_{\delta \mathbf{z}_{t}}^{i,j} -  \delta \mathbf{z}_t^{j}||_1.
\end{equation}
This computation of motion matrices with respect to the residual latents aids in modeling motion consistency. 
The motion sequence $\{\mathbf{M}_{\delta \mathbf{z}_{t}}^{i,j}\}_{t=1}^T$ is an input to DSS that facilitates the analysis of optimal transformation timesteps for frame $i$ and $j$. 
Additionally, this sequence aids in approximating the surrogate matrix used for transformations.


Given the intermediate step $\left( t^\ast \in [T] \right)$ switching from motion-based propagations to denoising (further details are provided in the subsequent section), the next task of MTN is to approximate the surrogate matrix $\mathcal{M}_{\mathbf{z}_t^\ast}^{i,j}$, by aggregating the motion dynamics captured within the denoising process from step $T$ to ${t^\ast}$. 
Given the consistency observed in motion dynamics throughout most diffusion steps, $\mathcal{M}_{\mathbf{z}_t^\ast}^{i,j}$ can be approximated by aggregating motion dynamics from step $T$ to step ${t^\ast}$. Using an MLP, $g_{\phi_3}$, this process is mathematically represented as:
\begin{equation}
    \mathcal{M}_{\mathbf{z}_t^\ast}^{i,j} = g_{\phi_3}(\mathbf{M}_{\delta \mathbf{z}_{t^\ast}}^{i,j}, \mathbf{M}_{\delta \mathbf{z}_{t^\ast+1}}^{i,j}, \ldots, \mathbf{M}_{\delta \mathbf{z}_{T}}^{i,j}).
\end{equation}

The second learning objective is to ensure accurate inter-frame transformations using the surrogate matrix, formulated as:
\begin{equation}
    \label{equ:transformation_error}
    \begin{aligned}
    \mathcal{L}^{\text{visual}}_\mathbf{z} 
    &= \sum_{i,j}|| \sum_{k={t^\ast}}^T ((\delta \mathbf{z}_k^{i})^\top \times \mathbf{M}_{\delta \mathbf{z}_{k}}^{i,j}) - \sum_{k={t^\ast}}^T \delta \mathbf{z}_k^{j}||_1 \\
    &\approx \sum_{i,j}|| (\sum_{k={t^\ast}}^T \delta \mathbf{z}_k^{i})^\top \times \mathcal{M}_{\mathbf{z}_t^\ast}^{i,j} - \sum_{k={t^\ast}}^T \delta \mathbf{z}_k^{j}||_1 
    \approx \sum_{i,j}||(\mathbf{z}_{t^\ast}^{i})^\top \times \mathcal{M}_{\mathbf{z}_{t^\ast}}^{i,j} - \mathbf{z}_{t^\ast}^{j}||_1.
    \end{aligned}
\end{equation}

The third learning objective involves ensuring temporal consistency and predicting future motion matrices. Specifically, the prediction process is formulated as using the sequence of observed motion matrices up to the last observed $R$-th frame to predict future $K$ motion matrices:
\begin{equation}
\label{equ:motion_prediction}
\hat{\mathcal{M}}_{\mathbf{z}_t^\ast}^{R, R+j} = g_{\phi_4}(\mathcal{M}_{\mathbf{z}_t^\ast}^{1, 2}, \mathcal{M}_{\mathbf{z}_t^\ast}^{2, 3}, \ldots, \mathcal{M}_{\mathbf{z}_t^\ast}^{R-1, R}), \quad \text{for } j \in [1, K],
\end{equation}
where $\phi_4$ represents the parameters of the motion prediction module. 
The prediction objective is the discrepancy between the predicted motion matrix and the ground truth motion matrix:
\begin{equation}
    \label{equ:prediction_error}
    \mathcal{L}^{\text{motion}}_\mathbf{z} = \sum_{j,t}||\hat{\mathcal{M}}_{\mathbf{z}_t^\ast}^{R, R+j} - \mathcal{M}_{\mathbf{z}_t^\ast}^{R, R+j}||_1.
\end{equation}
The prediction process helps maintain temporal consistency in the motion information and plays a vital role in enabling the generation of subsequent video frames with only a few reference frames. 

Therefore, the motion learning objective integrates the above three loss terms as follows:
\begin{equation}
    \label{equ:overall_loss}
    \mathcal{L}^\text{Trans} = \mathcal{L}^{\text{visual}}_{\delta \mathbf{z}} +  \mathcal{L}^{\text{visual}}_\mathbf{z} + \mathcal{L}^{\text{motion}}_\mathbf{z}.
\end{equation}

\subsection{Denoising Step Selector}
\label{sec:dss}
DSS is a meta-network designed to learn ${t^\ast}$, the proper intermediate step for switching from motion-based propagations to denoising. Specifically, the switch point  ${t^\ast}$ is determined to be the timestep which leads to the minimal weighted transformation error $\log (\beta \cdot t) \cdot e_t$, that is: 
\begin{equation}
     {t^\ast} = \argmin_{t\in \{1,\dots, T\}} \log (\beta \cdot t) \cdot e_t, 
\end{equation}
where $\beta$ is a hyperparameter balancing computational efficiency and transformation quality. Higher values of $\beta$ prioritize earlier denoising steps to enhance computation efficiency, whereas lower values focus on quality-preserving. 

To learn $t^\ast$, DSS takes the statistics derived from motion matrices $\{\mathbf{M}_{\delta \mathbf{z}_{t}}^{i,j}\}_{t=1}^T$ as input, including corresponding timestep indices and the NMI scores. 
It then implements a recurrent neural network~\cite{graves2012long} and outputs $\hat t$, the estimated most suitable switch step.
DSS is updated according to the cross-entropy loss between the predicted switching step $\hat t$ and the ground truth  $t^\ast$. 
During training, we apply a random mask to the input data to simulate scenarios with incomplete information. This strategy ensures that during inference, DSS does not require evaluation of the full sequence but can effectively optimize ${t^\ast}$ by analyzing only a subset of the available data, thereby reducing computational demands and speeding up the denoising process.

\begin{figure}[t]
    \centering
    \includegraphics[width=\textwidth]{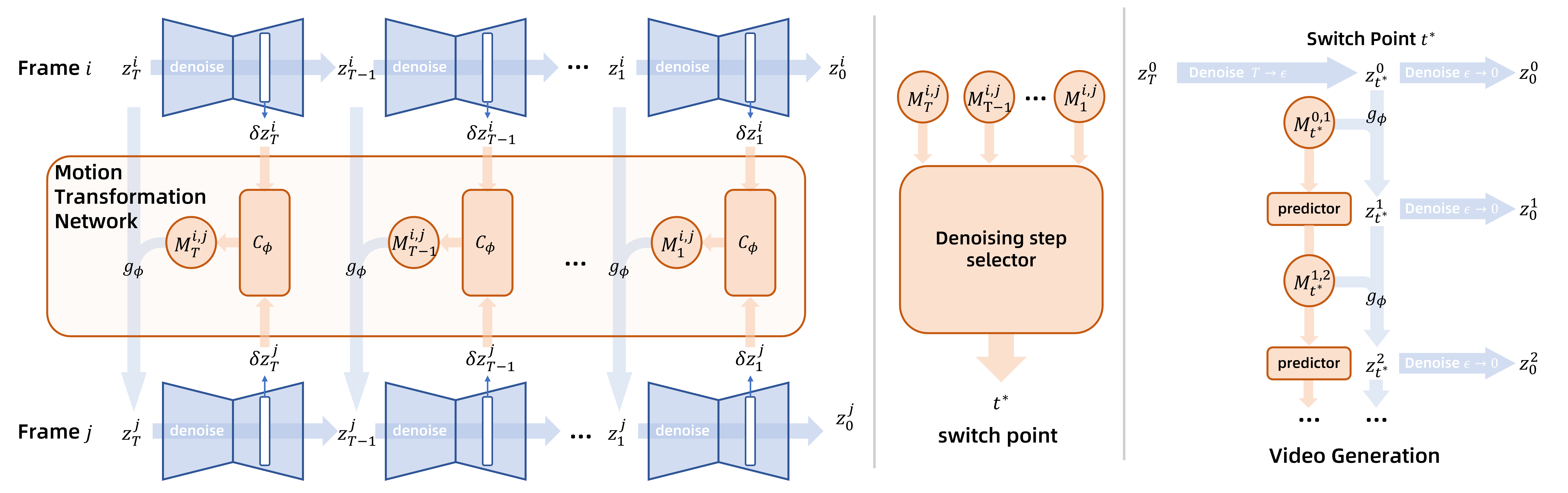}
\caption{Dr.~Mo consists of two main components: the Motion Transformation Network (MTN) and Denoising Step Selector (DSS). MTN learns motion matrices from semantic latents extracted from U-Net. 
The DSS is a meta-network that determines the appropriate transition step (denoted as $t^\ast$) for switching from motion-based propagations to denoising. After the transition step, those latent noise is processed by the rest of the diffusion model for video generation.}
\label{fig:model}
\end{figure}



%% file: 5_experiment.tex
\section{Experiments}
\label{sec:expri}

This section assesses Dr.~Mo's effectiveness in video generation and video editing. Additionally, we conduct ablation studies to explore the impact of varying denoising reuse strategies and to investigate the factors contributing to our method's capabilities.

\subsection{Implementation Details} 
\label{sec:implementation}
We use Stable Diffusion V1.5 \cite{rombach2022high} as the backbone, and train the proposed Dr.~Mo module on the WebVid-10M dataset~\citep{bain2021frozen}. We perform image resizing and center cropping to 512$\times$512, and downsample the video to \unit[4]{fps} to avoid low frame-to-frame variance. Training is conducted on the processed video with 20 consecutive frames randomly selected at a time. In this work, we use the representations from the block $\{6,8\}$ of the U-Net decoder to construct the motion matrix. (More hyperparameters can be found in the Appendix).

\subsection{Text-to-Video Generation}
\label{sec:quantitative}

We compare Dr.~Mo with several recent related works, including Latent-Shift~\citep{kim2024leveraging} and SimDA~\citep{xing2023simda}.
When comparing with other methods, we evaluate the zero-shot performance with text prompt from the test dataset of UCF-101~\citep{soomro2012ucf101} and MSR-VTT~\citep{xu2016msr}. 
For UCF-101, we write one template sentence for each class and utilize the sentence as a text prompt to generate 16 frames without fine-tuning. We report FVD~\cite{unterthiner2019fvd} and IS~\cite{salimans2016improved} on 10,000 samples following~\cite{ho2022video}. The generated samples have the same class distribution as the training set. 
For MSR-VTT, we report FID~\cite{heusel2017gans} and CLIPSIM~\cite{wu2021godiva} (average CLIP similarity between video frames and text), where all 2,990 captions from the test set are used, following~\cite{singer2022make}.

\begin{table}[t]
\centering
\caption{Comparison of video generation in terms of video quality and efficiency. }
\label{tab:quantitative}
\begin{tabular}{l>{\centering\arraybackslash}p{1.2cm}>{\centering\arraybackslash}p{1.1cm}>{\centering\arraybackslash}p{0.8cm}ccccccccc}
\toprule
Model          & Parameters & Tuned & \multicolumn{2}{c}{Speed(s)} & \multicolumn{2}{c}{UCF-101} & \multicolumn{2}{c}{MSR-VTT} \\ 
\cmidrule(lr){4-5} \cmidrule(lr){6-7} \cmidrule(lr){8-9}
               &                          &       & Res256 & Res512  & FVD$\downarrow$   & IS$\uparrow$    & FID$\downarrow$   & CLIPSIM$\uparrow$ \\ 
\midrule
Latent-Shift~\citep{kim2024leveraging}   & 1.53B                    & 0.880B  & 23.40     & -     & 360.04  & \textbf{92.72}   & 15.23   & 0.2773\\
Latent-VDM~\cite{kim2024leveraging}     & 1.58B                    & 0.920B  & 28.62     & -     & 358.34  & 90.74   & 14.35   & 0.2756\\
LVDM~\cite{he2022latent}           & 1.16B                    & 1.040B  & 21.23     & -     & 372.00  & -       & -       & 0.2930 \\
SimDA~\citep{xing2023simda}          & 1.08B                    & 0.025B  & 11.20     & 34.20 & -       & -       & -       & 0.2945\\
Video LDM~\cite{blattmann2023align}      & 4.20B                    & 2.650B  & -         & -     & 550.61  & 33.45   & -       & 0.2929\\
Make-A-Video~\cite{singer2022make}   & 9.72B                    & 9.720B  & -         & -     & 367.23  & 82.55   & 13.17   & 0.3049\\ \midrule
Dr.~Mo (Ours)           & 1.35B                    & 0.266B  & \textbf{6.57}      & \textbf{23.62} & \textbf{312.81}  & 89.63   & \textbf{12.38}   & \textbf{0.3056}\\ 
\bottomrule
\end{tabular}
\end{table}

\noindent \textbf{Quantitative Results.} 
As shown in Table~\ref{tab:quantitative}, Dr.~Mo outperforms competing video generation models, achieving the lowest FVD score of 312.81 on UCF-101 and the highest CLIPSIM score of 0.3056 on MSR-VTT. These results indicate that Dr.~Mo produces videos that closely match real videos in visual and temporal dynamics, and are semantically aligned with their corresponding inputs. Dr.~Mo differs from prior work primarily in its use of motion information and denoising step selection, and this is likely the cause of its superior performance.

\noindent \textbf{Qualitative Results.}
Figure~\ref{fig:qualitative} presents qualitative results for Dr.~Mo on the UCF-101 and MSR-VTT datasets. 256$\times$256 and 512$\times$512 resolution videos are considered. 
More examples can be found at our website~\footnote[1]{https://drmo-denoising-reuse.github.io/}. 

\noindent \textbf{Efficiency Evaluation.} 
As for the computing efficiency, Dr. Mo uses \unit[266]{M} of parameters and achieves the fastest reported inference rates, generating $16 \times$512$\times$512 frames in 23.62 seconds and generating 
16$\times$256$\times$256 frames in 6.57 seconds. This is notable considering some current models like those in Latent-Shift~\citep{an2023latentshift} only produce 256$\times$256 resolution images at similar parameter counts. These results suggest that Dr. Mo’s design, which optimizes the use of motion information, effectively reduces computational demands and speeds up video generation.

\begin{figure}[t]
    \centering
    \includegraphics[width=\linewidth]{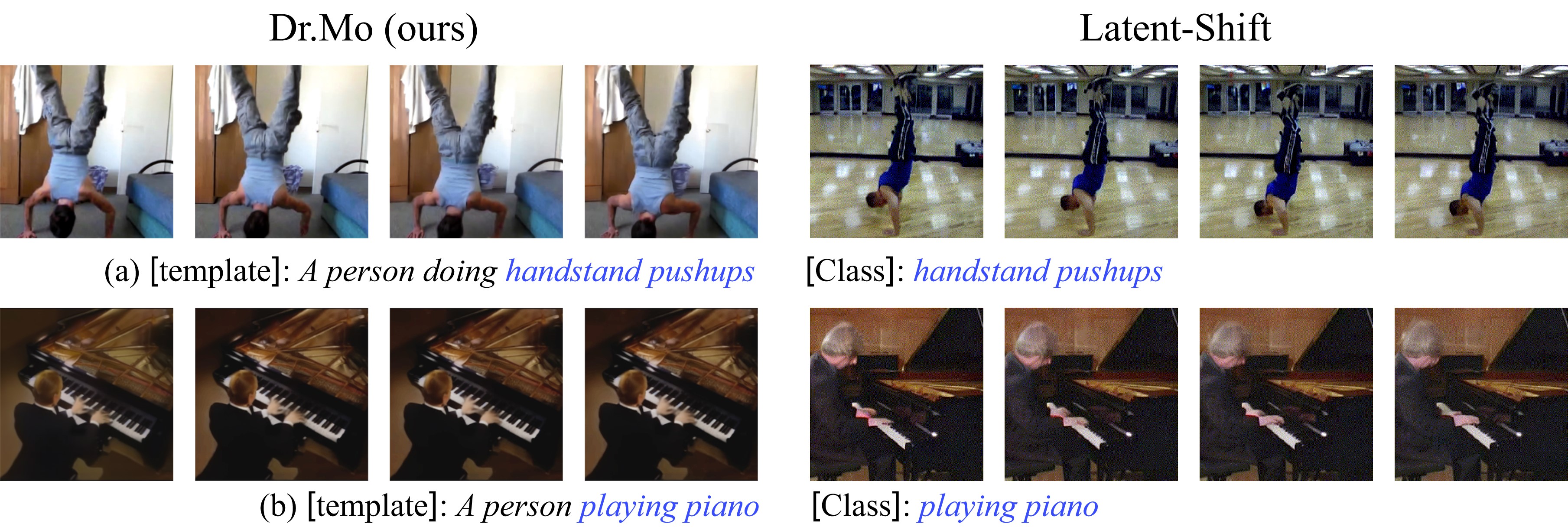}
    \vspace{-1em}
    \caption{
    Comparison with Latent-Shift using video frames with 256$\times$256 resolution on UCF-101.
    }
    \label{fig:qualitative}
\end{figure}


\begin{figure}[t]
    \centering
    \includegraphics[width=\textwidth]{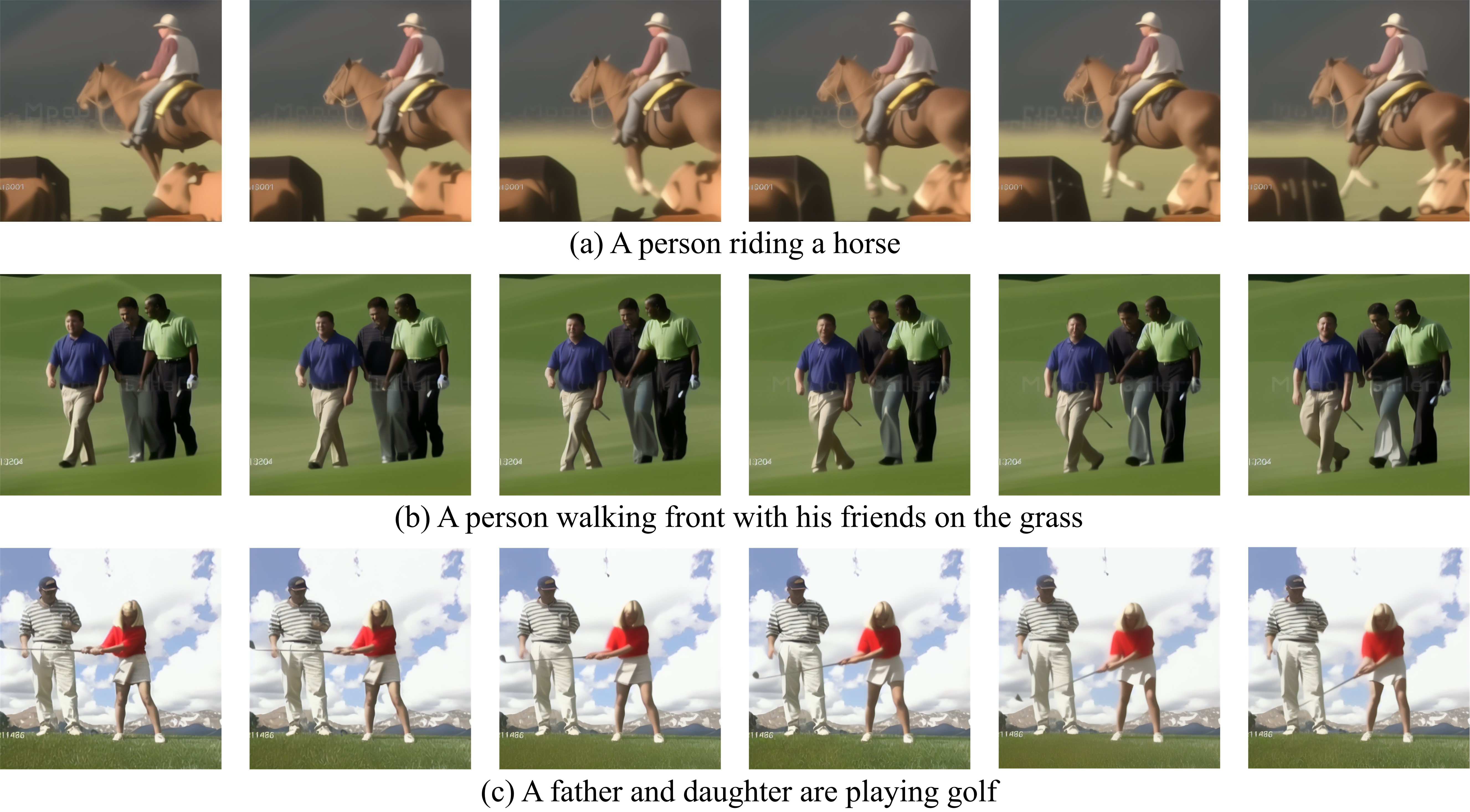}
    \vspace{-1em}
    \caption{Generated videos with 512$\times$512 resolution. }
    \label{fig:qualitative}
\end{figure}

\subsection{Video Editing}
\label{sec:video_editing}

\begin{figure}[t]
    \centering
    \includegraphics[width=\textwidth]{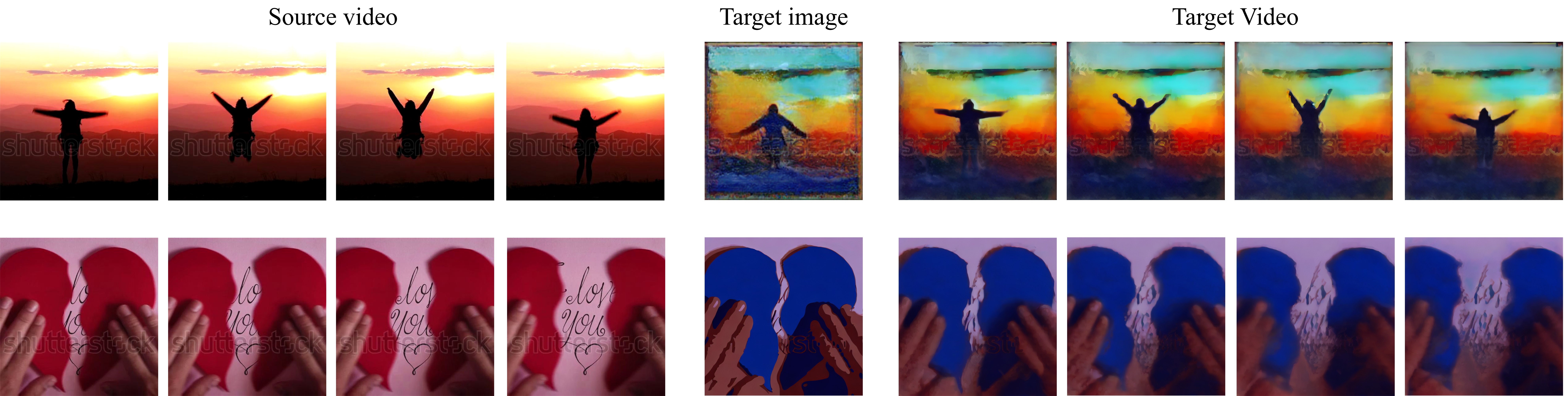}
    \vspace{-1em}
    \caption{Video editing Results.}
    \label{fig:editing}
\end{figure}

We evaluate Dr.~Mo's video editing capabilities by applying style transformations to real-world videos. 
Using the motion information from a reference video clip, we extract the motion matrix and apply it to the style transferred first frame to generate subsequent frames. 
As shown in Figure~\ref{fig:editing}, Dr.~Mo can transform real-world videos to match the visual style of the reference frame. 
Dr.~Mo learns to capture motion information, enabling it to produce stylistically diverse videos with realistic motion.

\begin{figure}[t]
    \centering
    \includegraphics[width=0.8\textwidth]{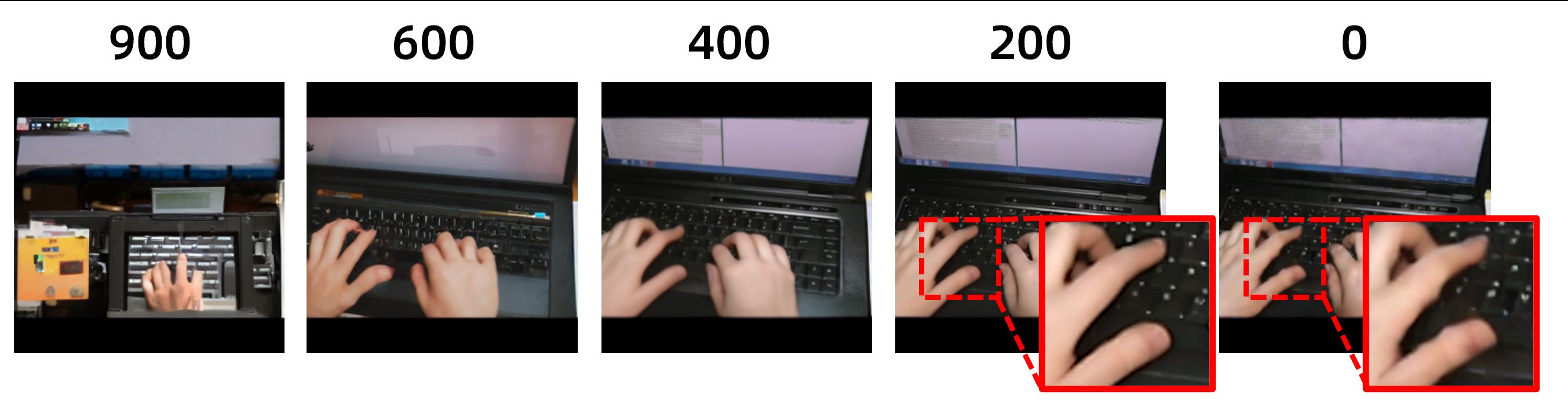}
    \caption{The result of motion transformation at different $t^\ast$ values. Too small $t^\ast$ will produce incorrect appearance details, while too large $t^\ast$ will lead to the destruction of visual features.}
    \label{fig:ablation1}
\end{figure}

\begin{figure}[h!]
    \centering
    \includegraphics[width=0.8\textwidth]{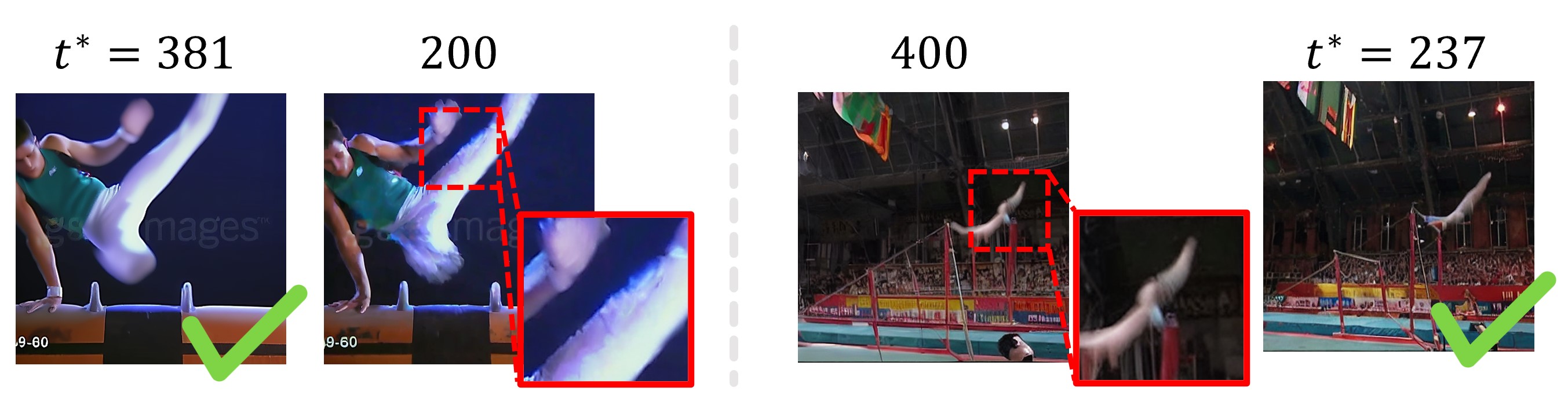}
    \caption{Left: Example of low motion consistency that requires a larger $t^\ast$ transformation. Right: Example of high motion consistency that requires a smaller $t^\ast$ transformation.}
    \label{fig:ablation2}
\end{figure}

\subsection{Ablation Study}

\noindent \textbf{Effect of Denoising Reuse.}
We conduct an ablation study to assess the impact of denoising reuse on video generation performance in Dr.~Mo by testing various switch points at steps $900$, $600$, $400$, $200$, and $1$. 
As shown in Figure~\ref{fig:ablation1}, Dr.~Mo performs optimally with $200$ denoising steps, indicating that an intermediate level of denoising provides the best balance between efficiency and video quality. 
At step $900$, excessive noises mask motion and visual features lead to ineffective transformations and compromised video content. Conversely, at step $1$, the presence of fine-grained visual features complicates motion modeling, resulting in accurate overall outlines but incorrect appearance details that degrade video quality.

\noindent \textbf{Effect of Varying Motion Consistency.}
We aim to assess the impact of varying motion consistencies on video generation. Following the methodology in MMVP~\citep{zhong2023mmvp}, we employ SSIM~\citep{wang2004image} as a metric and select two data samples with differing consistencies from WebVid. 
The left figure illustrates a video with low motion consistency, with the DSS predicting step 381 as optimal. Our results for steps 381 and 200 show that at step 200, there is a noticeable loss of detail information. Conversely, the right figure shows a video with high motion consistency; here, DSS identifies step 237 as optimal. While the results at step 237 are satisfactory, those at step 400 are less than ideal, due to insufficient learning of motion information. This is attributed to a deficiency in fine-grained visual features and inadequately learned related motion features. These observations highlight the crucial role of motion consistency over time and also validate the effectiveness of the DSS.



%% file: 4_related.tex
\section{Related Work}
\label{sec:related_work}
Recent advances in diffusion-based models~\cite{ho2022video,singer2022make,ho2022imagen,villegas2022phenaki} have produce high-quality videos by integrating spatio-temporal operations into traditional image-based frameworks. However, their reliance on iterative denoising processes makes them computationally expensive and unnecessarily slow. 
To simplify video generation, recent research has turned to latent space-based models~\citep{esser2023structure,xing2023simda,guo2023animatediff,wu2023tuneavideo}, particularly latent diffusion models~\citep{song2020denoising,ho2020denoising}. 
For instance, LVDM~\citep{he2022latent} and LaVie~\citep{wang2023lavie} generate sparse video patterns and interpolate intermediate latents, but do not explicitly model motion information. 
Latent-Shift~\citep{an2023latentshift} uses feature maps from adjacent frames to facilitate motion learning without extra parameters, while Text2Video-Zero~\citep{khachatryan2023text2video} employs predefined direction vectors to introduce motion dynamics, yet struggles with temporal consistency. 
VideoLCM~\citep{wang2023videolcm}  employs a teacher-student framework to distill consistency to minimize steps. However, it requires fine-tuning the complete diffusion process for each frame, taking 10s to generate 16$\times$256$\times$256 frames. In contrast, our approach takes only 6.57s with 200 steps using DDPM~\citep{ho2020denoising}. VidRD~\citep{gu2023reuse} also reuses latent features from previously generated clips does not adapt the number of reuse steps across frames, limiting its efficiency. 

To the best of our knowledge, this is the first work to study inter-frame motion consistency and use it to guide adaptive denoising reuse, significantly speeding up video generation.


%% file: 0_appendix.tex
\section{Hyperparameter Settings}
\label{sec:hyper}

For video data, we sample 20 frames from a four-second clip and train our model on 8 A100 GPUs. In the training step, in order to improve training efficiency, we first use low resolution for pre-training, in which we resize and center crop the image to 256$\times$256. 
Then moving to train our model on a high resolution, in which we resize and center crop the image to 512$\times$512. The following table describes the hyperparameters.

\begin{table}[htbp]
\centering
\caption{The hyper-parameter setting of our models.}
\label{table:hyperparameters}
\begin{tabular}{>{\centering\arraybackslash}p{6cm} >{\centering\arraybackslash}p{3cm} >{\centering\arraybackslash}p{3cm}}
\toprule
Hyper-parameter & low resolution & high resolution \\
\midrule
Image Size & 256 & 512 \\
Num Frame & 16  &  16 \\
Reference Frame &  4 & 4 \\
Latent Size & 32  & 64 \\
Guidance Scale &  7.5 & 7.5 \\
Text Encoder &  CLIPTextModel & CLIPTextModel \\
First Stage Model & AutoencoderKL  & AutoencoderKL \\
AE Out Channel &  4 &  4\\
Diffusion base Channel & 320  & 320 \\
Conditioning Key & crossattn & crossattn \\
Diffusion Step & 1000  & 1000 \\
Sample Scheduler & DDPM & DDPM \\
Num Res Block & 2  & 2 \\
Transformer Depth & 1  & 1 \\
Num Filter Layer & 3  & 3 \\
Base Filter Channel & 128  & 128 \\
Use Position Embedding &  TRUE & TRUE \\
Filter zero init &  TRUE & TRUE \\
Num Predictor Block & 10 & 10 \\
Base Predictor Channel &  256 & 256 \\
Learning Rate &  1e-4 & 3e-5 \\
Batch Size & 32 & 8 \\
Optimizer &  Adam & Adam \\
\bottomrule
\end{tabular}
\end{table}

\section{U-Net Block Analysis}

\label{sec:block_analysis}

\begin{figure}[h]
    \centering
    \includegraphics[width=\linewidth]{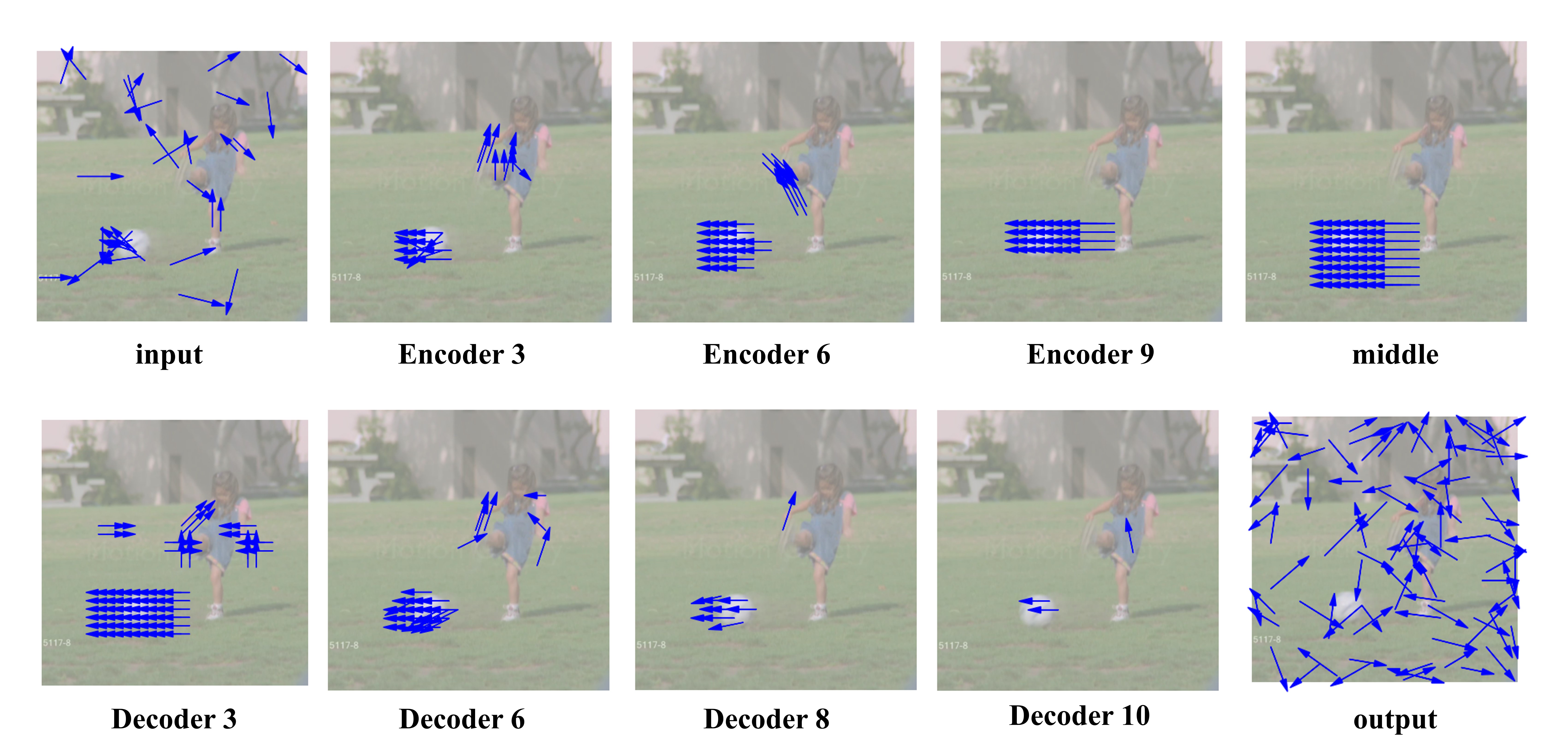}
    \caption{Visualization of transform matrix from different U-Net blocks.}
    \label{fig:block_analysis}
\end{figure}

By using the output features of each block of the pre-trained Stable Diffusion V1.5 model, we calculated and visualized the inter-frame transform matrix. The results showed that the features from U-Net middle layer could achieve a good transform matrix.
Ultimately, we select the coarse-grained layer decoder 6 (downsample 16)and the fine-grained layer decoder 8 (downsample 8), which both showed optimal performance. We combined the transform matrices from these two layers, and a voting network determined the type of transform applied to each feature.

\section{Limitations Discussions}
\label{sec:limitation}
In our current approach to generating longer videos or videos with larger motions, we have identified a limitation: the motion transformation process can result in a loss of visual information, leading to blurry outputs. In our future work, we will focus on complex motion scenarios or extended sequences. We aim to address this by exploring advanced motion modeling techniques and optimization strategies, enhancing both the fidelity and clarity of the generated videos.

\clearpage

%% file: neurips_2024.bbl
\begin{thebibliography}{35}
\providecommand{\natexlab}[1]{#1}
\providecommand{\url}[1]{\texttt{#1}}
\expandafter\ifx\csname urlstyle\endcsname\relax
  \providecommand{\doi}[1]{doi: #1}\else
  \providecommand{\doi}{doi: \begingroup \urlstyle{rm}\Url}\fi

\bibitem[An et~al.(2023)An, Zhang, Yang, Gupta, Huang, Luo, and Yin]{an2023latentshift}
Jie An, Songyang Zhang, Harry Yang, Sonal Gupta, Jia-Bin Huang, Jiebo Luo, and Xi~Yin.
\newblock Latent-shift: Latent diffusion with temporal shift for efficient text-to-video generation.
\newblock \emph{arXiv preprint arXiv:2304.08477}, 2023.

\bibitem[Bain et~al.(2021)Bain, Nagrani, Varol, and Zisserman]{bain2021frozen}
Max Bain, Arsha Nagrani, G{\"u}l Varol, and Andrew Zisserman.
\newblock Frozen in time: A joint video and image encoder for end-to-end retrieval.
\newblock In \emph{Proceedings of the IEEE/CVF International Conference on Computer Vision}, pp.\  1728--1738, 2021.

\bibitem[Baranchuk et~al.(2021)Baranchuk, Rubachev, Voynov, Khrulkov, and Babenko]{baranchuk2021label}
Dmitry Baranchuk, Ivan Rubachev, Andrey Voynov, Valentin Khrulkov, and Artem Babenko.
\newblock Label-efficient semantic segmentation with diffusion models.
\newblock \emph{arXiv preprint arXiv:2112.03126}, 2021.

\bibitem[Blattmann et~al.(2023)Blattmann, Rombach, Ling, Dockhorn, Kim, Fidler, and Kreis]{blattmann2023align}
Andreas Blattmann, Robin Rombach, Huan Ling, Tim Dockhorn, Seung~Wook Kim, Sanja Fidler, and Karsten Kreis.
\newblock Align your latents: High-resolution video synthesis with latent diffusion models.
\newblock In \emph{Proceedings of the IEEE/CVF Conference on Computer Vision and Pattern Recognition}, pp.\  22563--22575, 2023.

\bibitem[Esser et~al.(2023)Esser, Chiu, Atighehchian, Granskog, and Germanidis]{esser2023structure}
Patrick Esser, Johnathan Chiu, Parmida Atighehchian, Jonathan Granskog, and Anastasis Germanidis.
\newblock Structure and content-guided video synthesis with diffusion models.
\newblock In \emph{Proceedings of the IEEE/CVF International Conference on Computer Vision}, pp.\  7346--7356, 2023.

\bibitem[Graves \& Graves(2012)Graves and Graves]{graves2012long}
Alex Graves and Alex Graves.
\newblock Long short-term memory.
\newblock \emph{Supervised sequence labelling with recurrent neural networks}, pp.\  37--45, 2012.

\bibitem[Gu et~al.(2023)Gu, Wang, Zhao, Lu, Zhang, Wu, Xu, Zhang, Jiang, and Xu]{gu2023reuse}
Jiaxi Gu, Shicong Wang, Haoyu Zhao, Tianyi Lu, Xing Zhang, Zuxuan Wu, Songcen Xu, Wei Zhang, Yu-Gang Jiang, and Hang Xu.
\newblock Reuse and diffuse: Iterative denoising for text-to-video generation.
\newblock \emph{arXiv preprint arXiv:2309.03549}, 2023.

\bibitem[Guo et~al.(2023)Guo, Yang, Rao, Wang, Qiao, Lin, and Dai]{guo2023animatediff}
Yuwei Guo, Ceyuan Yang, Anyi Rao, Yaohui Wang, Yu~Qiao, Dahua Lin, and Bo~Dai.
\newblock Animatediff: Animate your personalized text-to-image diffusion models without specific tuning.
\newblock \emph{arXiv preprint arXiv:2307.04725}, 2023.

\bibitem[He et~al.(2022)He, Yang, Zhang, Shan, and Chen]{he2022latent}
Yingqing He, Tianyu Yang, Yong Zhang, Ying Shan, and Qifeng Chen.
\newblock Latent video diffusion models for high-fidelity long video generation.
\newblock \emph{arXiv preprint arXiv:2211.13221}, 2022.

\bibitem[Heusel et~al.(2017)Heusel, Ramsauer, Unterthiner, Nessler, and Hochreiter]{heusel2017gans}
Martin Heusel, Hubert Ramsauer, Thomas Unterthiner, Bernhard Nessler, and Sepp Hochreiter.
\newblock Gans trained by a two time-scale update rule converge to a local nash equilibrium.
\newblock \emph{Advances in neural information processing systems}, 30, 2017.

\bibitem[Ho et~al.(2020)Ho, Jain, and Abbeel]{ho2020denoising}
Jonathan Ho, Ajay Jain, and Pieter Abbeel.
\newblock Denoising diffusion probabilistic models.
\newblock \emph{Advances in neural information processing systems}, 33:\penalty0 6840--6851, 2020.

\bibitem[Ho et~al.(2022{\natexlab{a}})Ho, Chan, Saharia, Whang, Gao, Gritsenko, Kingma, Poole, Norouzi, Fleet, et~al.]{ho2022imagen}
Jonathan Ho, William Chan, Chitwan Saharia, Jay Whang, Ruiqi Gao, Alexey Gritsenko, Diederik~P Kingma, Ben Poole, Mohammad Norouzi, David~J Fleet, et~al.
\newblock Imagen video: High definition video generation with diffusion models.
\newblock \emph{arXiv preprint arXiv:2210.02303}, 2022{\natexlab{a}}.

\bibitem[Ho et~al.(2022{\natexlab{b}})Ho, Salimans, Gritsenko, Chan, Norouzi, and Fleet]{ho2022video}
Jonathan Ho, Tim Salimans, Alexey Gritsenko, William Chan, Mohammad Norouzi, and David~J Fleet.
\newblock Video diffusion models.
\newblock \emph{Advances in Neural Information Processing Systems}, 35:\penalty0 8633--8646, 2022{\natexlab{b}}.

\bibitem[Horn \& Schunck(1981)Horn and Schunck]{horn1981determining}
Berthold~KP Horn and Brian~G Schunck.
\newblock Determining optical flow.
\newblock \emph{Artificial intelligence}, 17\penalty0 (1-3):\penalty0 185--203, 1981.

\bibitem[Khachatryan et~al.(2023)Khachatryan, Movsisyan, Tadevosyan, Henschel, Wang, Navasardyan, and Shi]{khachatryan2023text2video}
Levon Khachatryan, Andranik Movsisyan, Vahram Tadevosyan, Roberto Henschel, Zhangyang Wang, Shant Navasardyan, and Humphrey Shi.
\newblock Text2video-zero: Text-to-image diffusion models are zero-shot video generators.
\newblock In \emph{Proceedings of the IEEE/CVF International Conference on Computer Vision}, pp.\  15954--15964, 2023.

\bibitem[Kim et~al.(2024)Kim, Jo, Jeon, Kim, Ahn, Kim, et~al.]{kim2024leveraging}
Yulhwa Kim, Dongwon Jo, Hyesung Jeon, Taesu Kim, Daehyun Ahn, Hyungjun Kim, et~al.
\newblock Leveraging early-stage robustness in diffusion models for efficient and high-quality image synthesis.
\newblock \emph{Advances in Neural Information Processing Systems}, 36, 2024.

\bibitem[Liu et~al.(2024)Liu, Xu, Yang, Zeng, and He]{liu2024drag}
Haofeng Liu, Chenshu Xu, Yifei Yang, Lihua Zeng, and Shengfeng He.
\newblock Drag your noise: Interactive point-based editing via diffusion semantic propagation.
\newblock \emph{arXiv preprint arXiv:2404.01050}, 2024.

\bibitem[Namekata et~al.(2024)Namekata, Sabour, Fidler, and Kim]{namekata2024emerdiff}
Koichi Namekata, Amirmojtaba Sabour, Sanja Fidler, and Seung~Wook Kim.
\newblock Emerdiff: Emerging pixel-level semantic knowledge in diffusion models.
\newblock \emph{arXiv preprint arXiv:2401.11739}, 2024.

\bibitem[Rombach et~al.(2022)Rombach, Blattmann, Lorenz, Esser, and Ommer]{rombach2022high}
Robin Rombach, Andreas Blattmann, Dominik Lorenz, Patrick Esser, and Bj{\"o}rn Ommer.
\newblock High-resolution image synthesis with latent diffusion models.
\newblock In \emph{Proceedings of the IEEE/CVF conference on computer vision and pattern recognition}, pp.\  10684--10695, 2022.

\bibitem[Ronneberger et~al.(2015)Ronneberger, Fischer, and Brox]{ronneberger2015u}
Olaf Ronneberger, Philipp Fischer, and Thomas Brox.
\newblock U-net: Convolutional networks for biomedical image segmentation.
\newblock In \emph{Medical image computing and computer-assisted intervention--MICCAI 2015: 18th international conference, Munich, Germany, October 5-9, 2015, proceedings, part III 18}, pp.\  234--241. Springer, 2015.

\bibitem[Salimans et~al.(2016)Salimans, Goodfellow, Zaremba, Cheung, Radford, and Chen]{salimans2016improved}
Tim Salimans, Ian Goodfellow, Wojciech Zaremba, Vicki Cheung, Alec Radford, and Xi~Chen.
\newblock Improved techniques for training gans.
\newblock \emph{Advances in neural information processing systems}, 29, 2016.

\bibitem[Singer et~al.(2022)Singer, Polyak, Hayes, Yin, An, Zhang, Hu, Yang, Ashual, Gafni, et~al.]{singer2022make}
Uriel Singer, Adam Polyak, Thomas Hayes, Xi~Yin, Jie An, Songyang Zhang, Qiyuan Hu, Harry Yang, Oron Ashual, Oran Gafni, et~al.
\newblock Make-a-video: Text-to-video generation without text-video data.
\newblock \emph{arXiv preprint arXiv:2209.14792}, 2022.

\bibitem[Song et~al.(2020)Song, Meng, and Ermon]{song2020denoising}
Jiaming Song, Chenlin Meng, and Stefano Ermon.
\newblock Denoising diffusion implicit models.
\newblock \emph{arXiv preprint arXiv:2010.02502}, 2020.

\bibitem[Soomro et~al.(2012)Soomro, Zamir, and Shah]{soomro2012ucf101}
Khurram Soomro, Amir~Roshan Zamir, and Mubarak Shah.
\newblock Ucf101: A dataset of 101 human actions classes from videos in the wild.
\newblock \emph{arXiv preprint arXiv:1212.0402}, 2012.

\bibitem[Unterthiner et~al.(2019)Unterthiner, van Steenkiste, Kurach, Marinier, Michalski, and Gelly]{unterthiner2019fvd}
Thomas Unterthiner, Sjoerd van Steenkiste, Karol Kurach, Rapha{\"e}l Marinier, Marcin Michalski, and Sylvain Gelly.
\newblock Fvd: A new metric for video generation.
\newblock 2019.

\bibitem[Villegas et~al.(2022)Villegas, Babaeizadeh, Kindermans, Moraldo, Zhang, Saffar, Castro, Kunze, and Erhan]{villegas2022phenaki}
Ruben Villegas, Mohammad Babaeizadeh, Pieter-Jan Kindermans, Hernan Moraldo, Han Zhang, Mohammad~Taghi Saffar, Santiago Castro, Julius Kunze, and Dumitru Erhan.
\newblock Phenaki: Variable length video generation from open domain textual descriptions.
\newblock In \emph{International Conference on Learning Representations}, 2022.

\bibitem[Wang et~al.(2023{\natexlab{a}})Wang, Zhang, Zhang, Liu, Zhang, Gao, and Sang]{wang2023videolcm}
Xiang Wang, Shiwei Zhang, Han Zhang, Yu~Liu, Yingya Zhang, Changxin Gao, and Nong Sang.
\newblock Videolcm: Video latent consistency model.
\newblock \emph{arXiv preprint arXiv:2312.09109}, 2023{\natexlab{a}}.

\bibitem[Wang et~al.(2023{\natexlab{b}})Wang, Chen, Ma, Zhou, Huang, Wang, Yang, He, Yu, Yang, et~al.]{wang2023lavie}
Yaohui Wang, Xinyuan Chen, Xin Ma, Shangchen Zhou, Ziqi Huang, Yi~Wang, Ceyuan Yang, Yinan He, Jiashuo Yu, Peiqing Yang, et~al.
\newblock Lavie: High-quality video generation with cascaded latent diffusion models.
\newblock \emph{arXiv preprint arXiv:2309.15103}, 2023{\natexlab{b}}.

\bibitem[Wang et~al.(2004)Wang, Bovik, Sheikh, and Simoncelli]{wang2004image}
Zhou Wang, Alan~C Bovik, Hamid~R Sheikh, and Eero~P Simoncelli.
\newblock Image quality assessment: from error visibility to structural similarity.
\newblock \emph{IEEE transactions on image processing}, 13\penalty0 (4):\penalty0 600--612, 2004.

\bibitem[Wu et~al.(2021)Wu, Huang, Zhang, Li, Ji, Yang, Sapiro, and Duan]{wu2021godiva}
Chenfei Wu, Lun Huang, Qianxi Zhang, Binyang Li, Lei Ji, Fan Yang, Guillermo Sapiro, and Nan Duan.
\newblock Godiva: Generating open-domain videos from natural descriptions.
\newblock \emph{arXiv preprint arXiv:2104.14806}, 2021.

\bibitem[Wu et~al.(2023)Wu, Ge, Wang, Lei, Gu, Shi, Hsu, Shan, Qie, and Shou]{wu2023tuneavideo}
Jay~Zhangjie Wu, Yixiao Ge, Xintao Wang, Stan~Weixian Lei, Yuchao Gu, Yufei Shi, Wynne Hsu, Ying Shan, Xiaohu Qie, and Mike~Zheng Shou.
\newblock Tune-a-video: One-shot tuning of image diffusion models for text-to-video generation.
\newblock In \emph{Proceedings of the IEEE/CVF International Conference on Computer Vision}, pp.\  7623--7633, 2023.

\bibitem[Xing et~al.(2023)Xing, Dai, Hu, Wu, and Jiang]{xing2023simda}
Zhen Xing, Qi~Dai, Han Hu, Zuxuan Wu, and Yu-Gang Jiang.
\newblock Simda: Simple diffusion adapter for efficient video generation.
\newblock \emph{arXiv preprint arXiv:2308.09710}, 2023.

\bibitem[Xu et~al.(2016)Xu, Mei, Yao, and Rui]{xu2016msr}
Jun Xu, Tao Mei, Ting Yao, and Yong Rui.
\newblock Msr-vtt: A large video description dataset for bridging video and language.
\newblock In \emph{Proceedings of the IEEE conference on computer vision and pattern recognition}, pp.\  5288--5296, 2016.

\bibitem[Zadaianchuk et~al.(2024)Zadaianchuk, Seitzer, and Martius]{zadaianchuk2024object}
Andrii Zadaianchuk, Maximilian Seitzer, and Georg Martius.
\newblock Object-centric learning for real-world videos by predicting temporal feature similarities.
\newblock \emph{Advances in Neural Information Processing Systems}, 36, 2024.

\bibitem[Zhong et~al.(2023)Zhong, Liang, Zharkov, and Neumann]{zhong2023mmvp}
Yiqi Zhong, Luming Liang, Ilya Zharkov, and Ulrich Neumann.
\newblock Mmvp: Motion-matrix-based video prediction.
\newblock In \emph{Proceedings of the IEEE/CVF International Conference on Computer Vision}, pp.\  4273--4283, 2023.

\end{thebibliography}
